\documentclass{article}
\usepackage{ijcai21}

\usepackage{times}

\usepackage{soul}
\usepackage{url}
\usepackage[hidelinks]{hyperref}
\usepackage[utf8]{inputenc}
\usepackage[small]{caption}
\usepackage{graphicx}
\usepackage{amsmath}
\usepackage{booktabs}
\usepackage{subfig}
\usepackage{amssymb}
\usepackage{amsfonts}
\title{Error-Robust Multi-View Clustering: Progress, Challenges and Opportunities}

\author{
Mehrnaz Najafi$^1$\footnote{Contact Author}\and
Lifang He$^2$\and
Philip S. Yu$^1$\\
\affiliations
$^1$Department of Computer Science, University of Illinois at Chicago, Chicago, IL, USA\\
$^2$Department of Computer Science \& Engineering, Lehigh University, Bethlehem, PA, USA\\
\emails
\{mnajaf2, psyu\}@uic.edu,
lih319@lehigh.edu
}

\begin{document}

\maketitle

\begin{abstract}
With recent advances in data collection from multiple sources, multi-view data has received significant attention. In multi-view data, each view represents a different perspective of data. Since label information is often expensive to acquire, multi-view clustering has gained growing interest, which aims to obtain better clustering solution by exploiting complementary and consistent information across all views rather than only using an individual view. Due to inevitable sensor failures, data in each view may contain error. Error often exhibits as noise or feature-specific corruptions or outliers. Multi-view data may contain any or combination of these error types. Blindly clustering multi-view data i.e., without considering possible error in view(s) could significantly degrade the performance. The goal of error-robust multi-view clustering is to obtain useful outcome even if the multi-view data is corrupted. Existing error-robust multi-view clustering approaches with explicit error removal formulation can be structured into five broad research categories - sparsity norm based approaches, graph based methods, subspace based learning approaches, deep learning based methods and hybrid approaches, this survey summarizes and reviews recent advances in error-robust clustering for multi-view data. Finally, we highlight the challenges and provide future research opportunities.
\end{abstract}

\section{Introduction}

\begin{figure}[t!]
        \centering
        \subfloat{
                \centering
                \includegraphics[width=0.24\linewidth]{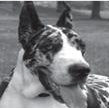}
                \label{fig:noise}\hfill}
        \subfloat{
                \centering
                \includegraphics[width=0.24\linewidth]{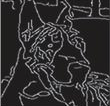}
                \label{fig:random-cor}\hfill}
        \subfloat{
                \centering
                \includegraphics[width=0.24\linewidth]{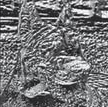}
                \label{fig:sample-cor}\hfill}
    \caption{Multi-View image. (a) Input image (color view) (b) Image edge (edge view) (c) Image texture (texture view) }
    \label{fig:multi-viewimage}
 \end{figure}

\begin{figure}[ht]
        \centering
         \includegraphics[width=\linewidth]{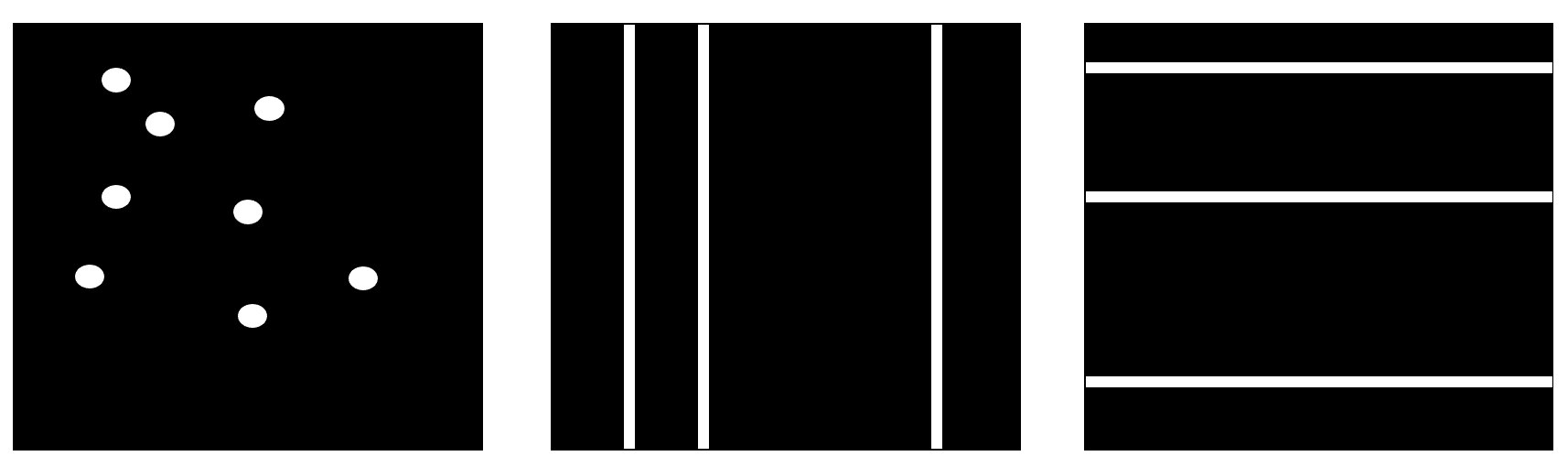}
        \caption{Three types of errors. (what we show is a corrupted data matrix whose rows are data instances and columns represent features). Left: Noise (or Random Corruptions), Middle: Feature-Specific Corruptions, Right: Sample-Specific Corruptions (or Outliers)}
        \label{fig:errors}
 \end{figure}

In the era of big data, data may often be provided by multiple views/perspectives, known as \textit{multi-view data}. For instance, an image may be available with different representations such as color, edge and texture where each representation is a view of data and describes a unique perspective of the same image. Figure \ref{fig:multi-viewimage} shows an example of multi-view data with three different views \cite{DBLP:journals/jifs/GoyalRS19}. The following properties are common in multi-view data:
\begin{itemize}
    \item Each view may encompass any type and number of features. Different views provide \textit{complementary} and \textit{consistent} information to each other \cite{zhao2017multi}. In addition, multi-view data may often come from diverse and multiple sources. 
    
    \item Due to inevitable system errors caused by sensor failures or malicious tampering, each view may be contaminated by \textit{error}. Error refers to deviation from clean assumptions (or assumed hypotheses) or presence of outlying data instances. 
    Three well-know types of errors are \textit{noise}, \textit{feature-specific corruptions}, or \textit{sample-specific corruptions (or outliers)}. Noise is defined as corruption of random subset of entries in data matrix where rows indicate data instances and columns refer to features. Feature-specific corruptions refer to perturbation of random subset of features, while sample-specific corruptions (or outliers) represent those data instances that are far away from the majority of other data instances in the feature space. Figure \ref{fig:errors} illustrates these error types. Multi-view data may contain any or combination of various error types \cite{DBLP:journals/jifs/GoyalRS19}. 
\end{itemize}

It is often not easy to collect label information for multi-view data in many applications. Lack of groundtruth to guide the learning process makes the underlying task much harder. \textit{Multi-view clustering} aims to find compatible clusters by exploiting complementary and consistent information among all views. 
Unfortunately, many existing multi-view clustering approaches assume that multi-view data does not contain error, which comprises their ability to produce promising clustering solution for erroneous multi-view data.

To handle erroneous multi-view data elegantly, several approaches have been proposed under \textit{error-robust multi-view clustering} where robustness signifies not too much sensitivity to error in data. The goal of error-robust multi-view clustering is to produce promising clustering solution even if the clean assumptions do not hold exactly or outliers are present.

Given the ever increasing interest towards clustering on erroneous multi-view data, we believe that a survey will help to guide the research to novel directions.
We distinguish two broad classes for error-robust multi-view clustering algorithms: (1) approaches that first remove error from possibly erroneous multi-view data using traditional error removal algorithms for single view \cite{10.5555/645924.671334,10.1145/335191.335388,1563979} and then use an existing multi-view clustering method that are not necessarily robust against error to obtain clustering solution across all views, (2) methods that propose specific error removal formulation for multi-view data. In other words, error removal and representation learning for clustering are performed simultaneously. 

Under the second class, we present five main error-robust multi-view clustering techniques: (1) Solely \textit{Sparsity norm based algorithms}, (2) \textit{Graph based methods}, (3) \textit{Subspace based learning methods}, (4) \textit{Deep learning based approaches} and (5) \textit{Hybrid algorithms}. Under the second class, the first category of approaches only impose sparsity norms such as $\ell_{2,1}$ on the representations to provide robustness against error. Graph based approaches construct pairwise similarity graph for each individual view and then employ multiple graphs fusion strategies to combine the graphs while constraining it with a term to encode error in the view. 

The third category of algorithms are built on the assumption that multi-view is drawn from (union of) subspace(s). Many datasets can be well characterized by subspace \cite{6618909,6180173,ZHU2018131}. These approaches first seek the underlying subspace structure for multi-view data and then derive clustering solution on the subspace representations. With error in multi-view data, views may not strictly follow the subspace structure. To provide robustness against error in subspace, error-robust subspace recovery has emerged that recovers the underlying clean subspace \cite{6180173,8425657}.
Deep learning based methods adopt recent trends in this area to conduct clustering on erroneous multi-view data. The last category of approaches, named as hybrid algorithms, are based on hybrid of two or more of the aforementioned categories to obtain clustering solution for multi-view data in a robust manner.

To this point, numerous surveys have been presented for multi-view clustering \cite{survey3,survey2,survey1}. However, to the best of our knowledge, none of the previous work have mainly focused on a survey on error-robust multi-view clustering methods. We therefore believe that this survey can provide concise and up-to-date knowledge about recent findings and developments in the field of error-robust multi-view clustering. The main contributions of this paper are as follows:
 
\begin{itemize}
    \item To the best of our knowledge, this survey is the first work on error-robust multi-view clustering. Most of the papers we surveyed have been published in well-known conferences in machine learning, artificial intelligence and data mining, e.g., NIPS, AAAI, IJCAI and ICDM. Journal papers were also included.
    
    \item We summarize and classify the state-of-the-art methods for error-robust multi-view clustering from a technical perspective. The provided information supports researchers to understand the past and current work in the field of error-robust clustering for multi-view data.
    
    \item We identify and discuss the technical challenges that need to be addressed together with suggestions for opportunities for future work.
\end{itemize}

\begin{figure*}
    \centering
    \includegraphics[width=0.8\textwidth,height=0.2\textheight]{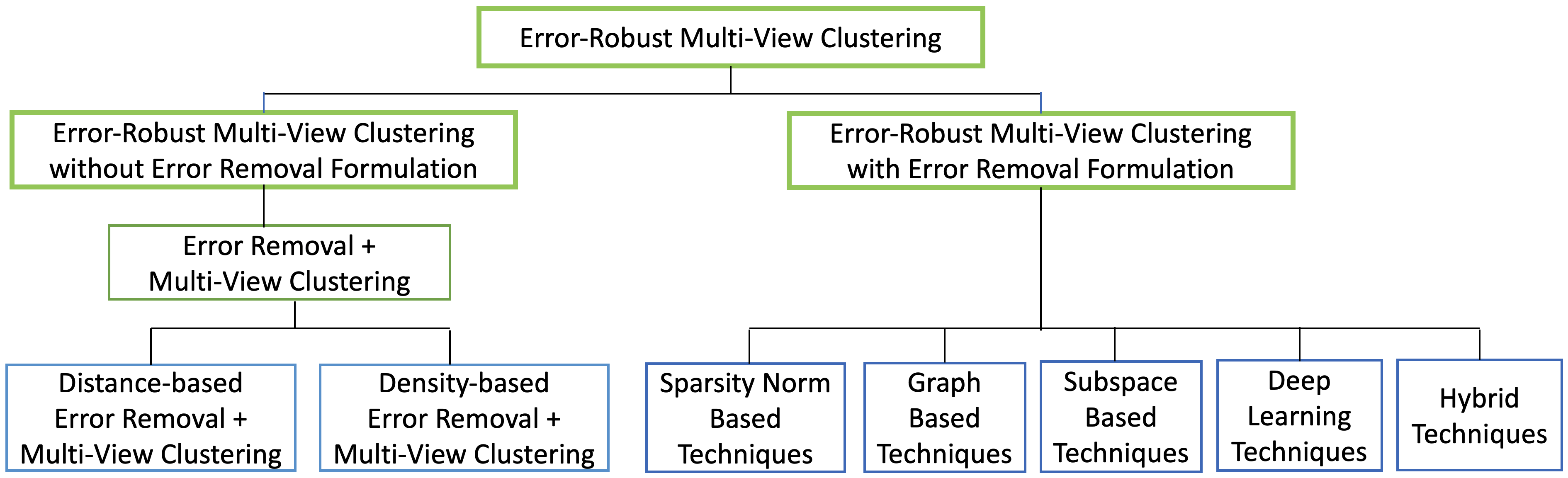}
    \caption{Taxonomy of error-robust multi-view clustering techniques}
    \label{fig:flowchart}
\end{figure*}

\begin{table*}[ht]
\scriptsize
\centering
\begin{tabular}{l|c}
\toprule
\textbf{Category} & \textbf{Publications}\\
\midrule
Sparsity norm methods & \cite{Cai2013MultiViewKC}, \cite{HUANG2018197}\\
\hline
Graph methods & \cite{Xia:2014:RMS:2892753.2892850}, \cite{DBLP:conf/bigdataconf/NajafiHY17}, \cite{ijcai2018-367}, \cite{LIANG2020220}, \cite{JING2021155}\\
\hline
Subspace methods & \cite{7298657}, \cite{7410839}, \cite{10.1109/ICCV.2015.185}, \cite{AAAIcss18}, \cite{DBLP:journals/ijcv/XieTZLZQ18}, \cite{pub.1110451961}, \cite{ijcai2019-404}\\
& \cite{9119408}, \cite{ZHANG2021324}\\
\hline
Deep learning methods & \cite{ijcai2019-409}\\
\hline
Hybrid methods & \cite{pmlr-v95-yu18a}\\
\bottomrule
\end{tabular}
\caption{Taxonomy and Representative Publications of Error-Robust Multi-View Clustering with Error Removal Formulation}
\label{tab:taxonomy}
\end{table*}

\section{Error-Robust Multi-View Clustering}
This section provides a technical overview of the recent research progress in error-robust clustering for multi-view data.

\textbf{Problem Description.} Given $n$ distinct data instances with $k$ possibly erroneous views denoted as $\mathbf{X}^{(1)} \in \mathbb{R}^{d_1 \times n}, \cdots, \mathbf{X}^{(k)} \in \mathbb{R}^{d_k \times n}$, the goal is to derive clustering solution with $c$ clusters across all views. We assume that all views provide complementary and consistent information to each other.  


\subsection{Error-Robust Multi-View Clustering with Traditional Error Removal for Single View}
The traditional approaches for error removal can be categorized into distance-based methods \cite{10.5555/645924.671334} and density-based method \cite{10.1145/335191.335388} that can be applied on the single-view. The distance-based approaches detect outliers on the single-view based on the distance of each data instances to its neighbours in feature space. For instance, Euclidean distance is one of the most widely-used metrics in two-dimensional feature space. The density-based techniques assign to each data instance in the single-view a degree of being outlier, which determines how isolated the data instance is compared to its neighbours.

\subsection{Error-Robust Multi-View Clustering with Error Removal Formulation for Multi-View}
In this section, we overview the recent techniques for error-robust multi-view clustering that explicitly address robustness against error in the proposed methods. In other words, they introduce specific terms in the proposed objective functions to provide robustness against error in multi-view data.
Table \ref{tab:taxonomy} provides taxonomy of existing methods under the category of error-robust multi-view clustering with error removal formulation.

\subsubsection{Solely Sparsity Norm Based Techniques}
One of the prominent approaches to provide robustness against error due to outliers is the $\ell_{2,1}$ norm. Cai et al. used this norm in order to adopt $k$-means for multi-view clustering and robustify the proposed method against outliers in data \cite{Cai2013MultiViewKC}. More precisely, the norm is imposed on the matrix factorization for each individual view which extracts view-specific and shared latent representation across all views. Another sparsity norm, named as capped norm, is used in \cite{HUANG2018197} to reduce the impact of error due to outliers on the clustering performance for multi-view $k$-means. The main drawback of this category of methods is that one specific norm can only deal with one type of error. In other words, a particular type of norm is not able to provide robustness against all error types simultaneously.

\subsubsection{Graph Based Techniques}
The importance of considering structure of data has been well recognized in the literature. Specifically, structure of data provides information about the relationships between data instances that can serve as a valuable guide for obtaining clustering solution.
Graph-based error-robust multi-view clustering algorithms focus on pairwise similarity graph where nodes indicate data instances and edge weights represent similarity between pairs of data instances. They leverage complementary and consistent information among multiple views by first mapping each individual view to the graph and then using multiple graph fusion strategy to derive clustering solution \cite{Xia:2014:RMS:2892753.2892850,DBLP:conf/bigdataconf/NajafiHY17,ijcai2018-367,LIANG2020220,JING2021155}. 

One of the most common graph representation approaches used for error-robust multi-view clustering is \textit{Markov chain}. Here, nodes represent data instances, while edge weights (or transitions) refer to normalized similarity between the corresponding data instances. Error in view can perturb transitions in Markov chain and thus leads to significant drop in the clustering performance. 

Xia et al. construct Markov chain for each individual view by using Gaussian kernel as similarity measure and then obtain shared Markov chain across all views \cite{Xia:2014:RMS:2892753.2892850}. The shared transition probability graph is optimized as a rank minimization problem while decomposing each individual view into true underlying view and error matrix that separates error from the corresponding view. The formulation only addresses noise. 
To capture various error types such as outliers and noise, Najafi et al. build shared Markov chain model across all views with a joint rank minimization, view decomposition and different structured sparsity inducing norms that encode common types of error elegantly \cite{DBLP:conf/bigdataconf/NajafiHY17}. 

Ren et al. established a novel graph-based method for robust multi-view clustering where each view is represented as a similarity graph \cite{ijcai2018-367}. The proposed method learns shared graph across all views while optimizing weight for each graph and enforcing regularization to encode noise in the graph. Liang et al. proposed a novel method that fuses the shared inter-view graph with multiple intra-view difference graphs, which aims to explicitly exploit the complementary information from intra-view difference graphs as well as further reduce the considerable noises in the shared graph \cite{LIANG2020220}.
Jing et al. proposed a novel approach for multi-view clustering which learns robust consistent common affinity graph across all views so as to minimize its disagreement with each individual view \cite{JING2021155}. In this method, each view is represented as a subspace of the corresponding Laplacian matrix. 

The existing graph-based techniques suffer from high computational complexity as they focus on pairwise similarity matrix construction which requires $O(n^2)$. Table \ref{tab:comparison-graph} summarizes the existing graph-based error-robust multi-view clustering approaches.

\begin{table*}[ht]
\scriptsize
\centering
\begin{tabular}{l|l|l}
\hline
\textbf{Method} & \textbf{Input} & \textbf{Graph Fusion Strategy}\\
\hline
\cite{Xia:2014:RMS:2892753.2892850} & Pairwise similarity graphs & learning shared latent transition probability matrix across all views\\
\hline
\cite{DBLP:conf/bigdataconf/NajafiHY17} & Pairwise similarity graphs & learning shared latent transition probability matrix across all views\\
\hline
\cite{ijcai2018-367} & Pairwise similarity graphs & learning shared latent similarity matrix\\
\hline
\cite{LIANG2020220} & Pairwise similarity graphs & learning shared latent transition probability matrix across all views\\
\hline
\cite{JING2021155} &  Laplacian of pairwise similarity graphs & learning shared latent Laplacian matrix\\
\hline
\end{tabular}
\caption{Comparison of Graph-based Error-Robust Multi-View Clustering Methods}
\label{tab:comparison-graph}
\end{table*}

\subsubsection{Subspace Based Techniques}
Broadly speaking, multi-view subspace clustering methods are built on the assumption that multi-view data is drawn from (union) of subspace(s) \cite{7298657,7410839,10.1109/ICCV.2015.185,AAAIcss18,DBLP:journals/ijcv/XieTZLZQ18,pub.1110451961,ijcai2019-404,ZHANG2019430,9119408,8691702,ZHANG2021324}. The challenging problem arises when data is erroneous because it may not strictly follow subspace structure. The solution is referred to as error-robust subspace recovery that recovers the underlying clean subspace for each individual view by decomposing it into two terms where the first term is the underlying clean subspace representation of the view and the second term captures error in the view as follows:

\begin{gather}
    \mathbf{X}^{(v)} = \mathbf{\tilde{X}}^{(v)} + \mathbf{E}^{(v)}
\end{gather}

\noindent where $\mathbf{\tilde{X}}^{(v)} \in \mathbb{R}^{d_v \times n}$ indicates the underlying clean subspace representation for view $v$ and $\mathbf{E}^{(v)} \in \mathbb{R}^{d_v \times n}$ denotes error in the view $v$.

All of the existing error-robust multi-view subspace clustering approaches utilize error-robust subspace recovery to provide robustness against error in multi-view data. Many of these techniques further use matrix factorization to approximate the underlying clean subspace representation term in the error-robust subspace recovery. Specifically, they seek the clean underlying subspace representation by conducting a distinct mapping $\mathbf{Z}^{(v)} \in \mathbb{R}^{n \times n}$ (subspace representation for view $v$) on the space of $\mathbf{X}^{(v)}$. This type of approximation is referred to as \textit{self-representation property} in the literature \cite{7410839,pub.1110451961,ijcai2019-404,7298657,AAAIcss18,9119408}. In this way, each
data instance is expressed by a linear combination of other data instances in the same subspace. The corresponding formulation is as follows:

\begin{gather}
\mathbf{X}^{(v)} = \mathbf{X}^{(v)} \mathbf{Z}^{(v)} + \mathbf{E}^{(v)} 
\end{gather}

The techniques based on self-representation property can be further categorized into two classes: (1) methods that construct affinity $\mathbf{Z} \in \mathbb{R}^{n \times n}$ across all $\mathbf{Z}^{(v)}$ which can be then given to any traditional clustering algorithm such as $k$-means to derive the final clustering solution \cite{7298657,AAAIcss18,9119408}, (2) approaches that jointly learns affinity matrix $\mathbf{Z}$ and derive clustering solution for all views \cite{7410839}.
Gao et al. formulate the clustering problem with a joint view-specific subspace representation learning and cluster indicator matrix optimization based on affinity matrix $\mathbf{Z}$ \cite{7410839}. The view-specific subspace representation learning is achieved using self-representation property. 


Under the second class of approaches, Cao et al. proposed a novel method that follows the same strategy, while explores complementary information across all views by focusing on diversity between views \cite{7298657}. 
To exploit specificity and consistency properties for better clustering solution, Luo et al. decompose underlying subspace representations in the self-representation property related constraint into two terms that encode these two properties \cite{AAAIcss18}. Zheng et al. developed a novel method that learns subspace reconstruction coefficients using multi-kernel learning and block diagonal regularizer to fully capture complementary information across multiple views \cite{9119408}. 

In order to explore high-order correlations underlying multi-view data for more comprehensive learning, a variety of existing error-robust multi-view subspace clustering methods use \textit{tensor} \cite{10.1109/ICCV.2015.185,DBLP:journals/ijcv/XieTZLZQ18}. Tensor is higher order generalization of vector (first order tensor) and matrix (second order tensor). More precisely, existing error-robust tensorized multi-view subspace clustering methods establishe a tensor by stacking subspace representations of all views and then formulate the problem as tensor rank minimization. They then build affinity matrix based on the learned subspace representations. The robustness is achieved by using self-representation property for each individual view. The drawback here lies on computational complexity. In other words, most of the existing tensorized algorithms have high computational complexity. Table \ref{tab:comparison-subspace} summarizes existing error-robust multi-view subspace clustering approaches.

\begin{table*}[ht]
\scriptsize
\centering
\begin{tabular}{l|l|l}
\hline
\textbf{Method} & \textbf{Type of Correlations} & \textbf{Subspace Fusion Strategy}\\
\hline
\cite{7410839} & pairwise correlations & affinity subspace computation during learning\\
\hline
\cite{7298657,AAAIcss18,9119408} & pairwise correlations & affinity subspace computation after learning\\
\hline
\cite{10.1109/ICCV.2015.185,DBLP:journals/ijcv/XieTZLZQ18} & high-order correlations & affinity subspace computation after learning\\
\hline
\end{tabular}
\caption{Comparison of Subspace-based Error-Robust Multi-View Clustering Methods}
\label{tab:comparison-subspace}
\end{table*}

\subsubsection{Deep Learning Based Techniques}
Auto-Encoders (AE) and generative adversarial network (GAN) are prominent components in deep learning that can provide robustness against error in multi-view data. To better capture non-linearity in multi-view data, deep learning based models for error-robust multi-view clustering have been received significant attention \cite{ijcai2019-409}. Li et al. combine three different loss functions named as autoencoder loss, GAN loss and clustering loss based on KL-divergence to obtain clustering solution among multiple views \cite{ijcai2019-409}. The proposed model learns latent representations across all views using (denoising) autoencoders. The autoencoders may produce blurred reconstructed representations. Thus, the proposed model employs GAN to further improve reconstructed results.


\subsubsection{Hybrid Techniques} 
Hybrid style methods for error-robust multi-view clustering combine several styles to construct clustering solution among all views \cite{pmlr-v95-yu18a}. Yu et al. devised a novel two stage process for error-robust multi-view clustering based on graph and subspace learning. In the first stage, the proposed model builds separate learners for each distinct view and then minimizes disagreement between view pairs while using self-representation property for robustness \cite{pmlr-v95-yu18a}. The learner construction term for each individual view is based on pairwise similarity graph regularization. The focus of the second stage is on the clustering solution construction based on learners.

\section{Challenges and Opportunities}
Although there have been quite a few developments for error-robust clustering on multi-view data in recent years, there are still issues that require more effective and efficient solutions as well as challenges that remain unresolved or underexplored. These challenges provide opportunities for future research on this area. In this section, we discuss broad challenges and opportunities the community faces.

\subsection{Challenge I: Prior Information}
Absence of label information to guide the learning process makes the underlying task of robust clustering much harder. It is therefore crucial to incorporate \textit{prior information} such as \textit{prior pairwise constraints} that describe the relationship between data instances into the clustering process. Specifically, prior pairwise constraints reveal information about structure of data that serves as a valuable guide for the learning process.

One of the most prominent types of prior pairwise constraints are \textit{must-link} and \textit{cannot-link} constraints. The must-link constraints specifies that a pair of data instances must belong to the same cluster, whereas cannot-link constraints enforces that a pair of data instances should be assigned to different clusters.
The challenge arises is then on jointly take advantage of multiple views and prior information for error-robust multi-view clustering. 

\textit{Opportunities.} To this point, Zhao et al. established a novel two step method, named as MVMC, to work around this challenge to some extent by constructing pairwise similarity matrix for each individual view based on the given pairwise constraints and learning the shared similarity matrix across all views by minimizing its distance with the similarity matrix of each individual view  \cite{10.1007/978-3-319-57529-2_32}. In the second step, any traditional clustering algorithm can be applied on the shared similarity matrix to obtain the final clustering solution. However, MVMC assumes that multi-view data does not encompass errors. 
None of the existing methods leverage prior information for error-robust multi-view clustering in a joint manner. Thus, a comprehensive solution to this problem is yet to be found.

\subsection{Challenge II: Online Multi-View Data}
In the era of big data, the size of multi-view data could be extremely large. Thus, it is not realistic to apply multi-view clustering on large datasets without considering computational complexity and memory requirement in an online fashion. The challenge is on how to combine multiple views with memory limitation and computational complexity requirement for error-robust multi-view clustering.

\textit{Opportunities.} To date, an effective and efficient solution for online multi-view clustering is to formulate the problem as a joint weighted non-negative matrix factorization problem \cite{7840701,DBLP:conf/aaai/HuC19}. To overcome memory requirement, the proposed methods process multi-view data chunk by chunk. However, they do not handle various error types in multi-view data elegantly. Therefore, a complete solution for this challenge is yet to be developed. 

\subsection{Challenge III: Incomplete View(s)}
In reality, in addition to error, due to nature of data or data collection cost, data instances within view(s) may be missing. View(s) with missing data instances are referred to as \textit{incomplete view(s)} \cite{7840701}. Clustering on data with incomplete view(s) is referred to as \textit{partial multi-view clustering}. The challenge involved in this direction is a comprehensive clustering algorithm for multiple views where view(s) may be both incomplete and erroneous.


\textit{Opportunities.} To this end, several approaches have been developed \cite{Rai_multiviewclustering,6729618,10.5555/2892753.2892826,10.1007/978-3-319-23528-8_20,7840701,8594983}. In particular, most of the existing approaches deal with incompleteness by either applying multiple kernel learning or non-negative matrix factorization. Besides, one of the recent deep learning work invokes GAN to generate missing data instances in incomplete views while learning the shared representation for clustering \cite{8594983}. A few of the other existing algorithms developed a two stage process: (1) missing multi-view data completion and (2) clustering on complete views. Although they have shown success for partial multi-view clustering, they neglect possible error in view(s). Furthermore, it has been shown that joint view completion and clustering often lead to better clustering solution. Therefore, there are more works to be done before we have a complete and full error-robust clustering method for data with incomplete view(s). 

\subsection{Challenge IV: View(s) with Missing Feature Values}
Besides various types of errors, each individual view may suffer from missing feature values. Traditionally, multi-view clustering methods have relied on the following ways to alleviate this problem to some extent: (1) filling out the missing feature values with feature mean computed based on the observed feature values, (2) using the most probable value for the missing feature values and (3) filling out the missing feature values with a global constant such as Unknown. However, they often do not yield promising results. The challenge is then how to fuse multiple views when they contain error and missing feature values.

\textit{Opportunities.} To this end, in addition to the aforementioned traditional approaches, Tao and Yu devised a multi step method that first fills out missing feature values and then conducts clustering \cite{DBLP2020}. The key issue of this approach is that it does not handle erroneous view elegantly. An effective solution thus needs to be developed.

\subsection{Challenge V: Multi-Task Learning}
\textit{Multi-task multi-view clustering} refers to performing multiple related tasks together and utilizing the relationship between these tasks to enhance clustering solution for multi-view data. An example of multi-task multi-view learning is to cluster business customer of different online shopping websites where clients are described with multiple views. The challenge is how to discover effective and efficient mechanism to perform multiple related tasks and exploit relationship between these tasks for more comprehensive learning under a unified framework \cite{survey1}.

\textit{Opportunities.} To this point, quite a few methods for multi-task multi-view clustering has been proposed \cite{ZHANG2012465,AlStouhi2014MultiTaskCU,7555376,ZHANG2019776}. Specifically, they utilize subspace structure or matrix factorization to model intra-task (within task) and domain adaptation for inter-task (between task) relationships. The domain adaptation here refers to knowledge transfer between different tasks. Unfortunately, all of the existing approaches assume that views are not erroneous. This calls for advanced techniques that exploit complementary and consistent information from multiple views while considering error in view(s) for error-robust multi-task multi-view clustering.

\subsection{Challenge VI: Deep Error-Robust Multi-View Clustering}
Although the community has started to use deep learning for error-robust multi-view clustering, learning high-quality latent representations for erroneous multi-view data is still under explored \cite{ijcai2019-409}. This defines a last challenge for the field of error-robust multi-view clustering.

\textit{Opportunities.} To this date, to the best of our knowledge, the method in \cite{ijcai2019-409} is the only deep error-robust multi-view clustering that explicitly separates error from view(s). The future directions of research include: (1) the clean underlying representations for erroneous multi-view should be learned and then clustering is conducted. Specifically, Zhou and Paffenroth have developed a novel variant of autoencoder which is robust against different types of error \cite{10.1145/3097983.3098052}. This idea can be utilized to propose a deep learning model that learns error-robust representations and obtains clustering solution jointly and (2) self-representation property in the loss function should be enforced to capture error in multi-view data.
\section{Conclusion}
Nowadays, data may encompass any type of error such as noise and outliers. Unfortunately, traditional multi-view clustering methods conduct clustering blindly i.e., without distinguishing error from view(s). This leads to unreliable and low-quality clustering solution. Since error is inevitable in multi-view data, it is of high importance to use error-robust clustering method for possibly erroneous multi-view data. The focus of this survey is to present progress, challenges and opportunities for error-robust multi-view clustering. More precisely, we reviewed technical trends and current state-of-the-art in algorithm development for error-robust multi-view clustering. 

To the best of our knowledge, this is the first work with focus on error-robust multi-view clustering. We also discussed vital challenges as well as opportunities that need to be addressed in this area of research. 
We believe that the challenges we described can be utilized as guidelines for the next generation of error-robust multi-view clustering. Given the fact that many methods have been developed so far, there are still lots of opportunities that yet need to be explored for mature development in this ever-evolving area of research.

\bibliographystyle{named}
\bibliography{ijcai21}
\end{document}